\documentclass[usletter]{article}
\usepackage{spconf}
\usepackage{rotating}
\usepackage{float}
\usepackage{booktabs}
\usepackage{arydshln}
\usepackage{subfig}
\restylefloat{table}
\usepackage{graphicx}
\usepackage[update]{epstopdf}
\usepackage{multirow}
\usepackage{hhline}

\newcommand{\mywidth}{0.5\textwidth}

\title{Resolution limits on visual speech recognition}

\name{{\em Helen L. Bear, Richard Harvey, Barry-John Theobald, Yuxuan Lan}}

\address{School of Computing Sciences , University of East Anglia, Norwich, NR4 7TJ, UK.\\
{\small \tt helen.bear@uea.ac.uk, r.w.harvey@uea.ac.uk, b.theobald@uea.ac.uk, y.lan@uea.ac.uk}}

\begin{document}
\maketitle

\begin{abstract}
Visual-only speech recognition is dependent upon a number of factors that can be difficult to control, such as: lighting; identity; motion; emotion and expression. But some factors, such as video resolution are controllable, so it is surprising that there is not yet a systematic study of the effect of resolution on lip-reading. Here we use a new data set, the Rosetta Raven data, to train and test recognizers so we can measure the affect of video resolution on recognition accuracy. We conclude that, contrary to common practice, resolution need not be that great for automatic lip-reading. However it is highly unlikely that automatic lip-reading can work reliably when the distance between the bottom of the lower lip and the top of the upper lip is less than four pixels at rest. 
\end{abstract}

\section{Introduction}
A typical lip-reading system has a number of stages: first, the data are pre-processed and normalised; second, the face and lips are tracked; third, visual features are extracted and classified. In practice many systems find tracking challenging, which affects the overall recognition performance. However, the tracking problem is not insurmountable and it is now realistic to track talking heads in outdoor scenes filmed with shaky hand-held cameras~\cite{Bowden:2012fk}, so we focus on feature extraction using Active Appearance Models (AAMs) \cite{AAMs}. We select AAMs since they have been shown to have robust performance on a number of datasets (~\cite{Matthews_Baker_2004,982900,ong2008robust,ong2009robust} for example) and out perform other feature types~\cite{lan2010improving}. 

\section{Dataset and feature extraction}

An AAM is a combined model of shape and appearance trained to fit to a whole video sequence \cite{AAMs}. Training creates a mean model and a set of modes, which may be varied to create shape and appearance changes. In training, a small number of frames are identified and manually landmarked. These models are Procrustes-aligned and the mean and covariance of the shape are computed. The eigenvectors of the covariance matrix give a set of modes of variation, which are used to deform the mean shape. For appearance a mesh shape-normalizes the images via a piecewise affine transform so the pixels of all images are aligned. We then compute the mean and the eigenvectors of their covariance. Concatenating the shape and appearance features forms the feature vector for training and testing. Having built a model on a few frames, it is fitted to unseen data using inverse compositional fitting \cite{Matthews_Baker_2004}.

\begin{figure*}[!ht]	
\centering
\setlength{\tabcolsep}{1pt}
	\begin{tabular}{c c c c} 
	
	\includegraphics[width=0.245\textwidth]{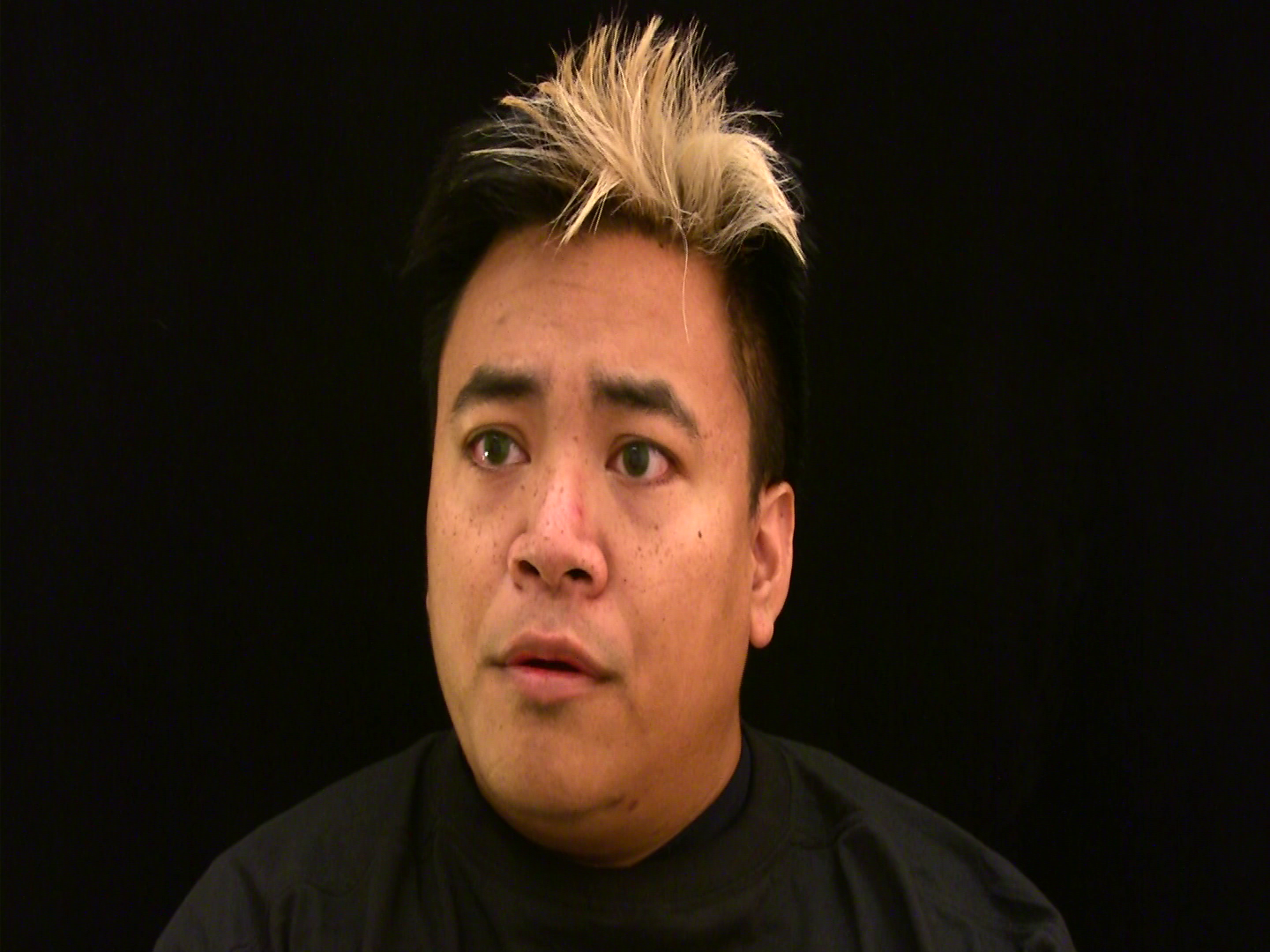} & 
	\includegraphics[width=0.245\textwidth]{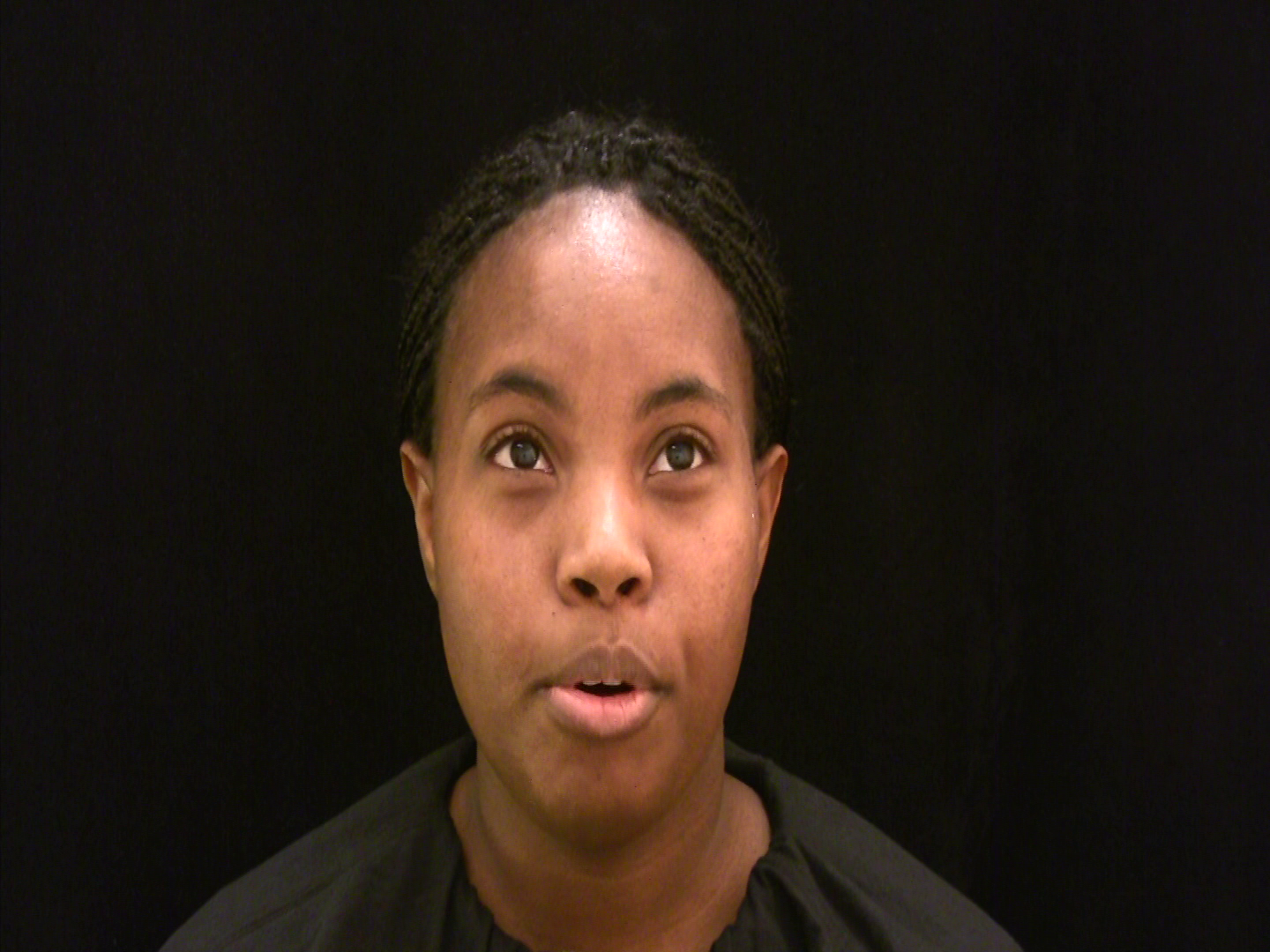} & 
	\includegraphics[width=0.245\textwidth]{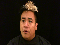} &
	\includegraphics[width=0.245\textwidth]{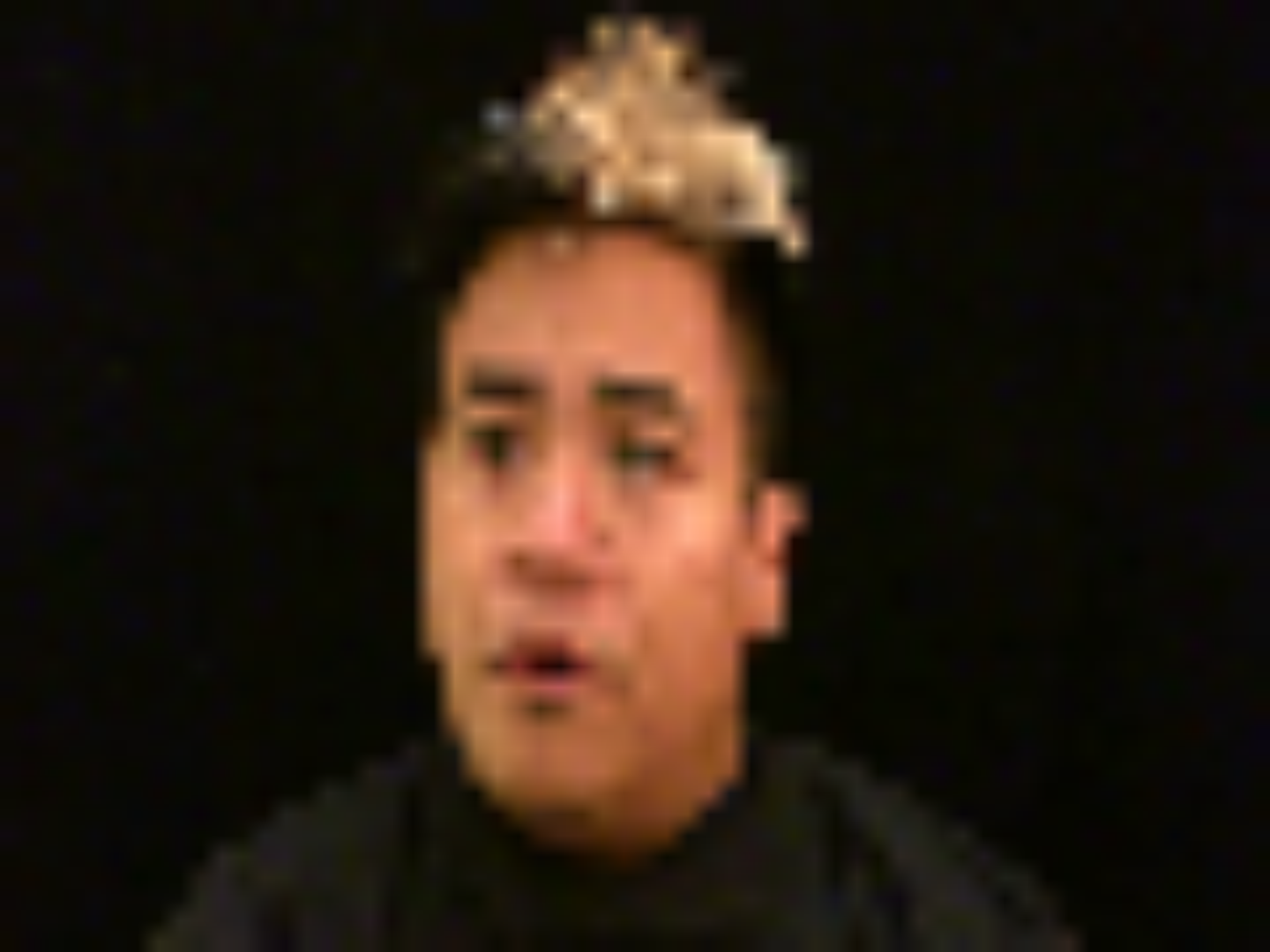} \\
	\multicolumn{2}{c}{(a)}&
	(b) &
	(c)
	\end{tabular}
	\caption{(a) $1440\times1080$-Original resolution image for T1 \& T2, (b) $60\times45$-T1 downsampled, and (c) $1440\times1080$-T1 restored}
	\label{tab:htkacckey}
\end{figure*}

The Rosetta Raven data are four videos of two North American talkers (each talker in two videos), reciting Edgar Allen Poe's `The Raven'. The poem was published in 1845 and, recited properly, the poem has trochaic octameter \cite{quinn1980critical}, but this does not appear to have been followed by the talkers in this dataset. Figure~\ref{tab:htkacckey}(a) shows example frames from the high-definition video of the two talkers. The database summarised in Table~\ref{tab:frames} was recorded at $1440\times 1080$ non-interlaced resolution at 60 frames per second. The talkers wore no make-up. 

\begin{table}[h]	
\centering
\caption{Frame images from each video}
	\begin{tabular}{| l | r | r | r |} 
	\hline
	Video & Train Images & Fit Images & Duration\\
	\hline
	Talker1 - 1 & 10 & 21,648 & 00:06:01 \\
	Talker1 - 2 & 10 & 21,703 & 00:06:02 \\
	Talker2 - 1 & 11 & 31,858 & 00:08:52 \\
	Talker2 - 2 & 11 & 33,328 & 00:09:17 \\
	\hline
	\end{tabular}
	\label{tab:frames}
\end{table} 

All four videos were converted into a set of images (one per frame) with ffmpeg using image2 encoding at full high-definition resolution ($1440\times1080$). 

\begin{figure}[!ht]
\centering
	\includegraphics[width=0.2\textwidth]{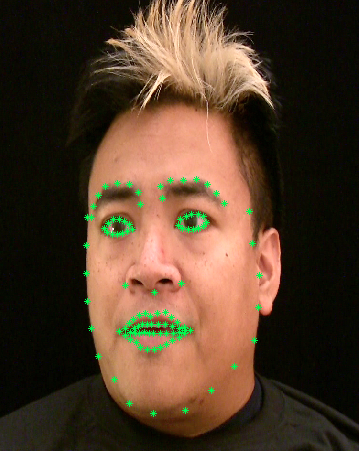} 
	\includegraphics[width=0.2\textwidth]{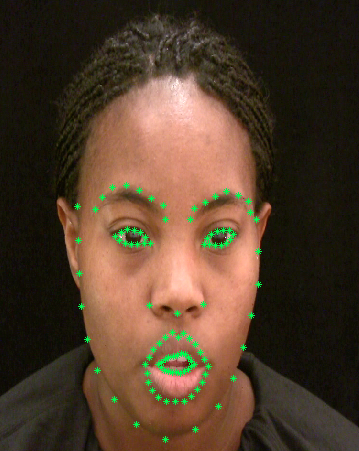} 
\caption{Showing full face mesh for talker T1 (left) and T2 (right)}
\label{fig:mesh}
\end{figure}

To build an initial model we select the first frame and nine or ten others randomly. These \emph{key frames} are hand-labelled with a model of a face and lips. This preliminary model is then fitted, via inverse compositional fitting~\cite{Matthews_Baker_2004} to the remaining frames (Table~\ref{tab:frames} lists total frames for each video). At this stage therefore we have tracked and fitted full face talker dependent AAMs on full resolution lossless PNG frame images as in Figure~\ref{fig:mesh}. 

\begin{figure}[!ht]
\centering
	\includegraphics[width=0.2\textwidth]{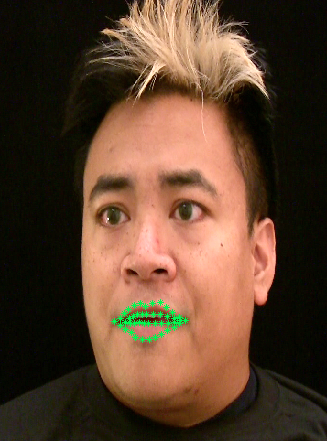} 
	\includegraphics[width=0.2\textwidth]{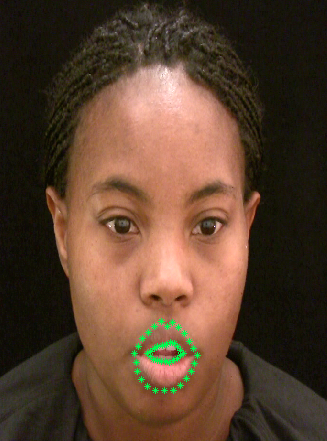} 
\caption{Showing lip-only mesh for talker T1 (left) and talker T2 (right)}
\label{fig:lip_models}
\end{figure}

These models are then decomposed into sub-models for the eyes, eyebrows, nose, face outline and lips (this allows us to obtain a robust fit from the full face model but process only the lips). Figure~\ref{fig:lip_models} shows both talker's lips sub-model. Next, the video frames used in the high-resolution fitting were down-sampled to each of the required resolutions (Table~\ref{table:resolution_sizes}) by nearest neighbor sampling and then up-sampled via bilinear sampling (Figure~\ref{tab:htkacckey}) to provide us with 18 sets of frames. These new frames are the same physical size as the original ($1440\times1080$) but contain far less information due to the downsampling.

\begin{table}[!ht]	
\centering
	\caption{Resolutions}
	\begin{tabular}{| l | l | l | l |} 
	\hline
	$1440\times1080$ & $960\times720$ & $720\times540$ &$ 360\times270$ \\
	 $240\times180$ & $180\times135$ & $144\times108$ & $120\times90$ \\
	 $90\times67$ & $80\times60$ & $72\times54$ & $65\times49$ \\ 
	$69\times45$ & $55\times42$ & $51\times39$ & $48\times36$ \\
	 $45\times34$ & $42\times32$ & & \\
	 \hline
	\end{tabular}
	\label{table:resolution_sizes}
\end{table}

We are most interested in the affect of low resolution on the loss of lip-reading information rather than, say the affect it would also have on the tracker (many AAM trackers lose track quite easily at low resolutions and we do not wish to be overwhelmed with catastrophic errors due to tracking problems which can often be solved in other ways~\cite{5459283}). Consequently the shape features in this experiment are unaffected by the downsample whereas as the appearance features vary (a useful benchmark as it will turn out).

For talker1 (T1), we retain 6 shape and 14 appearance parameters and for talker2 (T2), 7 shape and 14 appearance parameters. The number of parameters was chosen to retain 95\% of the variance in the usual way~\cite{AAMs}. 

\section{Recognition Method}

\begin{table}[h]
\centering
	\caption{Phone to viseme mapping}
		\begin{tabular} {| ll|ll |}
		\hline
		vID & Phones & vID & Phones \\
		\hline
		v01 & /p/ /b/ /m/ & v10 & /i/ /ih/ \\
		v02 & /f/ /v/ & v11 & /eh/ /ae/ /ey/ /ay/\\
		v03 & /th /dh/ & v12 & /aa/ /ao/ /ah/ \\
		v04 & /t/ /d/ /n/ /k/ /g/ /h/ /j/ & v13 & /uh/ /er/ /ax/ \\
		& /ng/ /y/  & & \\
		v05 & /s/ /z/ & v14 & /u/ /uw/ \\
		v06 & /l/ & v15 & /oy/ \\
		v07 & /r/ & v16 & /iy/ /hh/ \\ 
		v08 & /sh/ /zh/ /ch/ /jh/ & v17 & /aw/ /ow/ \\
		v09 & /w/ & v18 & silence \\
		\hline
		\end{tabular}
		\label{tab:visememapping}
\end{table}

\begin{figure}[!ht]
\centering
	\includegraphics[width=0.45\textwidth]{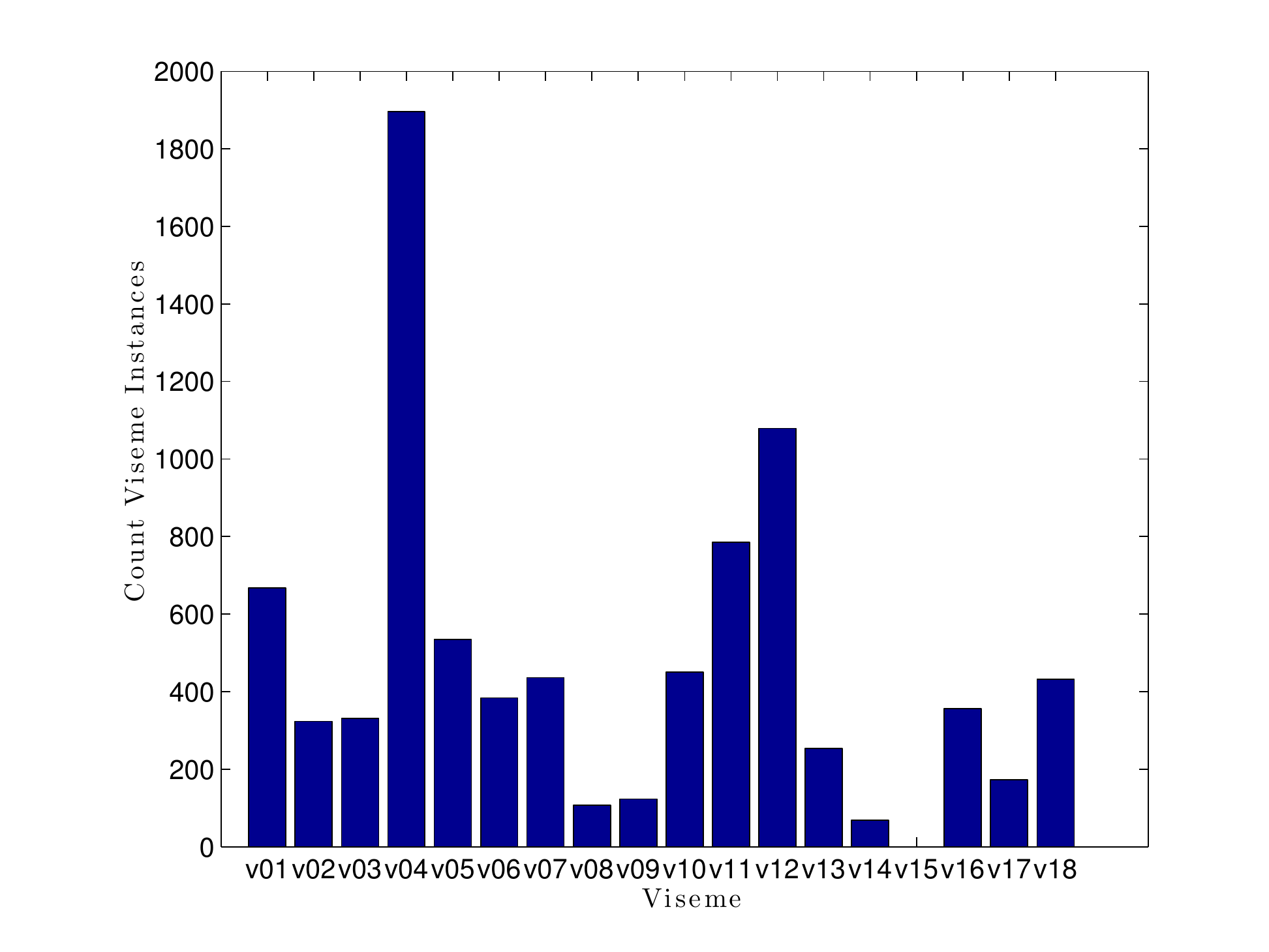} 
	\caption{Visemes present in both T1 videos}
	\label{fig:k1_vis_hist}
	\includegraphics[width=0.45\textwidth]{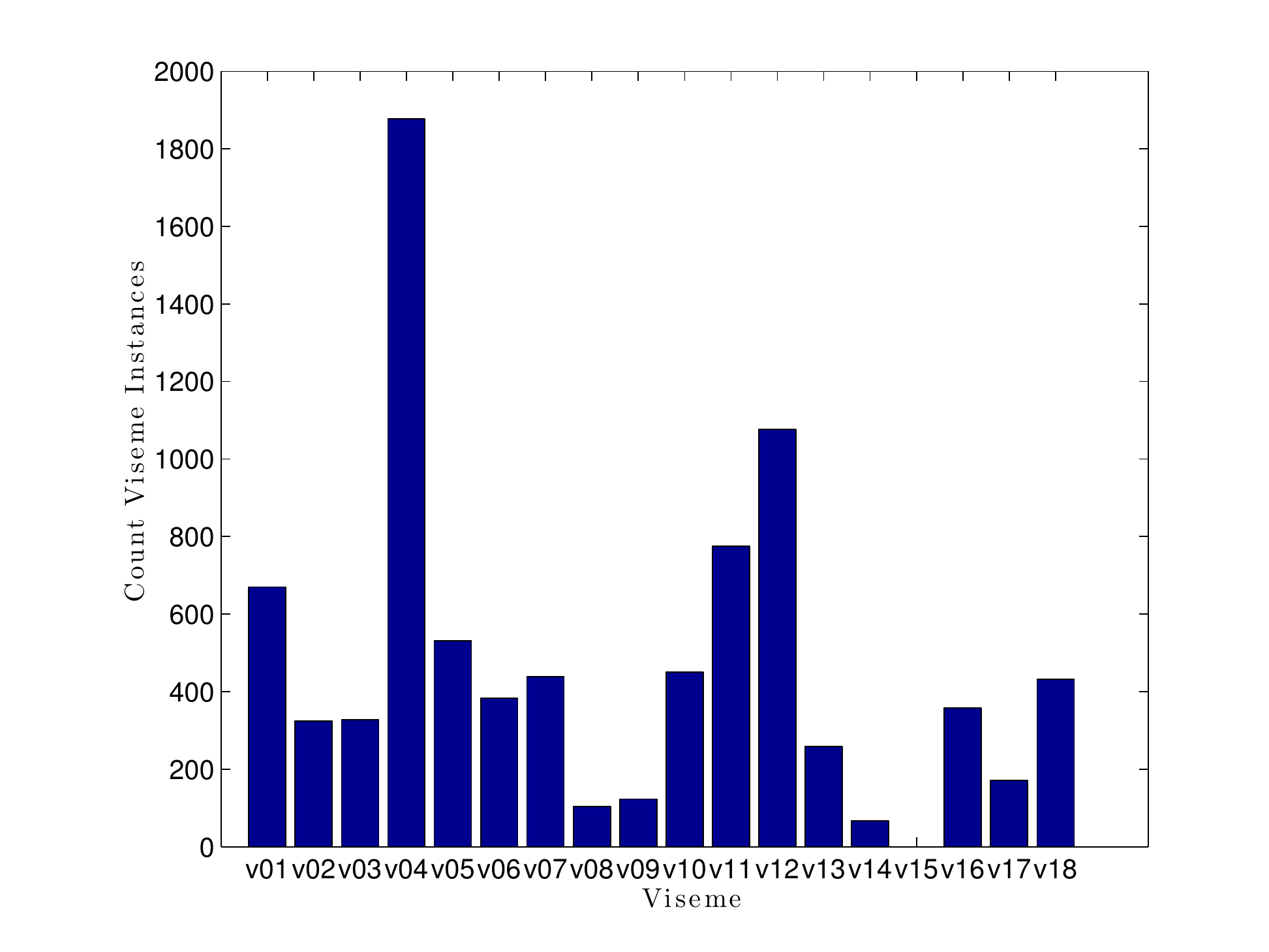} 
	\caption{Visemes present in both T2 videos}
	\label{fig:c1_vis_hist}
\end{figure}

To produce the ground truth we listen to each recitation of the poem and produced a ground truth text (some recitations of the poem were not word-perfect). This word transcript is converted to an American English phone level transcript using the CMU pronunciation dictionary \cite{cmudict}. However not all phones are visible on the lips, so we select a mapping from phones to \emph{visemes} (which are the visual equivalent of phonemes). Here, the viseme mapping is based upon Walden's trained consonants \cite{walden1977effects} and Montgomery et al's vowel \cite{massaro98:talking} classifications as illustrated in Table~\ref{tab:visememapping}. Viseme recognition is selected over phoneme recognition as, on a small data set, it has the benefits of reducing the number of classes needed (the model for each class forms a single recogniser) and increasing the training data available for each viseme classifier. Note that not all visemes are equally represented in the data as is shown by the viseme counts in Figures~\ref{fig:k1_vis_hist} and~\ref{fig:c1_vis_hist}.

For each talker, a test fold is randomly selected as 42 of the 108 lines in the poem. The remaining lines are used as training folds. Repeating this five times gives five-fold cross-validation. Visemes cannot be equally represented in all folds.

For recognition we use Hidden Markov Models (HMMs) implemented in the Hidden Markov Toolkit (HTK)~\cite{htk34}.
An HMM is initialised using the `flat start' method using a prototype of five states and five mixture components and the information in the training samples. We choose five states and five mixtures via \cite{982900}. We define an HMM for each viseme plus silence and short-pause labels (Table~\ref{tab:visememapping}) and re-estimate the parameters four times with no pruning.
We use the HTK tool \texttt{HHEd} to tie together the short-pause and silence models between states two and three before re-estimating the HMMs a further two times. 
Then \texttt{HVite} is used to force-align the data using the word transcript~\footnote{We use the \texttt{-m} flag with \texttt{HVite} with the manual creation of a viseme version of the CMU dictionary for word to viseme mapping so that the force-alignment produced uses the break points of the words.}. 

The HMMs are now re-estimated twice more, however now we use the force-aligned viseme transcript rather than the original viseme transcript used in the previous HMM re-estimations. 
To complete recognition using our HMMs we require a word network. We use \texttt{HLStats} and \texttt{HBuild} to make both a Unigram Word-level Network (UWN) and a Bi-gram Word-level Network (BWN). 
Finally \texttt{HVite} is used with the different network support for the recognition task and \texttt{HResults} gives us the correctness and accuracy values.

\section{Results}
Recognition performance of the HMMs can be measured by both correctness, $C$, and accuracy, $A$,
\[
C = \displaystyle \frac{N-D-S}{N},\quad
A = \displaystyle \frac{N-D-S-I}{N},
\]
where $S$ is the number of substitution errors, $D$ is the number of deletion errors, $I$ is the number of insertion errors and $N$ the total number of labels in the reference transcriptions~\cite{htk34}. 

We use accuracy as a measure rather than correctness since it accounts for all errors including insertion errors which are notoriously common in lip reading. An insertion error occurs when the recognizer output has extra words/visemes missing from the original transcript \cite{htk34}. As an example one could say ``Once upon a midnight dreary'', but the recognizer outputs ``Once upon upon midnight dreary dreary". Here the recognizer has inserted two words which were never present and it has deleted one.

\begin{figure}[!ht]
\centering
\includegraphics[width=\mywidth,keepaspectratio]{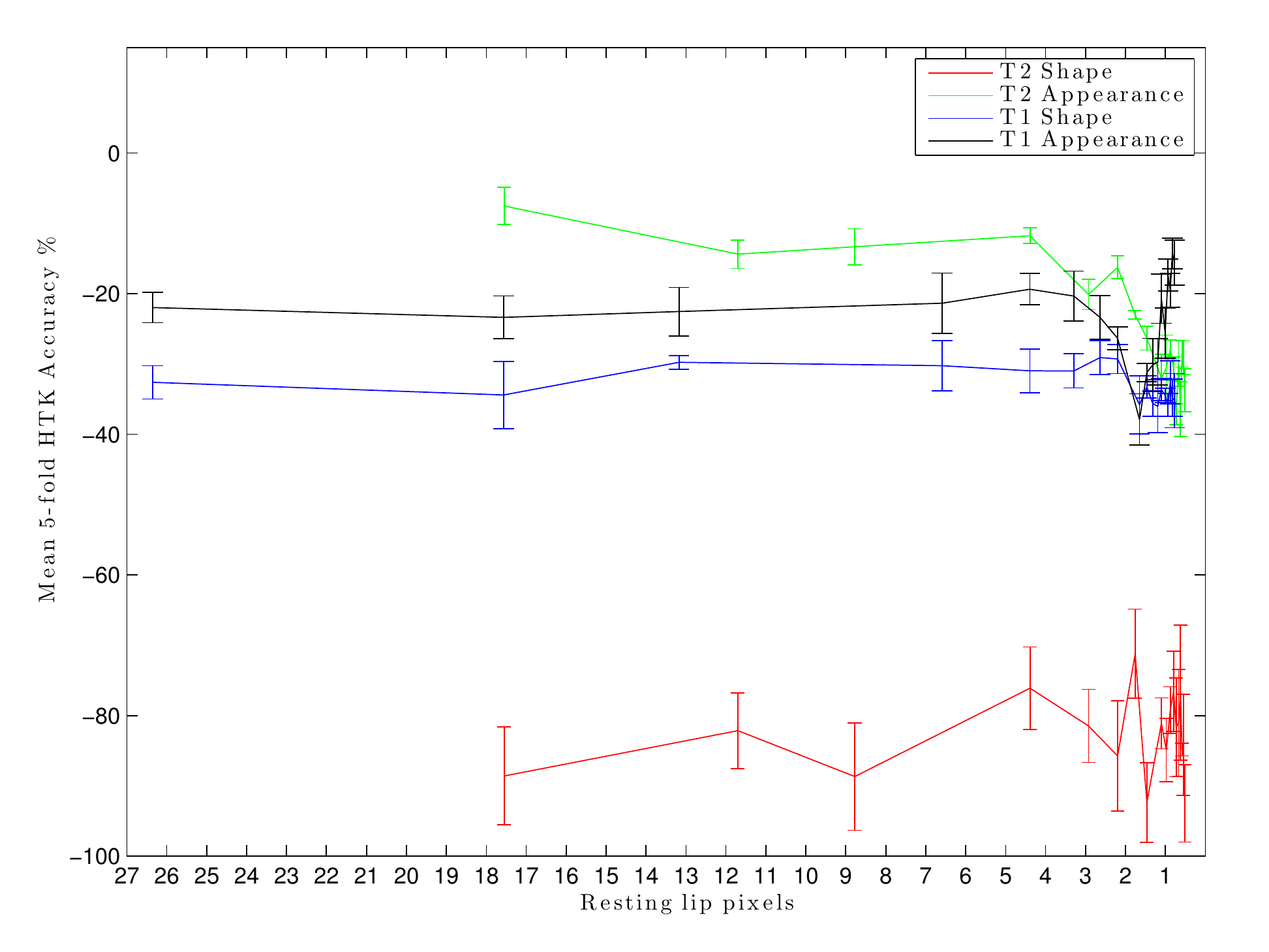} 
\caption{Mean viseme recognition accuracy supported by UWN at 18 degraded resolutions shown by vertical resting lip height in pixels. Error bars show $\pm$ one standard error.}
\label{fig:HTKAccUEPS}
\end{figure}

Figure~\ref{fig:HTKAccUEPS} shows the acurracy, $A$, versus resolution for an UWN. The $x$-axis is calibrated by the vertical height of the lips of each talker in their rest position. For example, at the maximum resolution of $1440\times1080$ talker T1 has a lip-height of approximately 26 pixels in the rest position whereas T2 has a lip-height of approximately 17 pixels. The worst performance is from talker T2 using shape-only features. Note that the shape features do not vary with resolution so any variation in this curve is due to the cross-fold validation error (all folds do not contain all visemes equally). Nevertheless the variation is within an error bar. The poor performance is, as usual with lip-reading, a standard error dominated by insertion errors (hence the negative $A$ values). The usual explanation for this effect is that shape data contains a few characteristic shapes (which are easily recognised) in a sea of indistinct shapes - it is easier for a recogniser to insert garbage symbols than it is to learn the duration of a symbol which has indistinct start and end shapes due to co-articulation. Talker T1 has more distinctive shapes so scores better on the shape feature.

However it is the appearance that is of more interest since this varies as we downsample. At resolutions lower than four pixels it is difficult to be confident that the shape information is effective. However the basic problem is a very low error rate (shown in Figure~\ref{fig:HTKAccUEPS}) therefore we adopt a more supportive word model.

\begin{figure}[!ht]
\centering
\includegraphics[width=\mywidth,keepaspectratio]{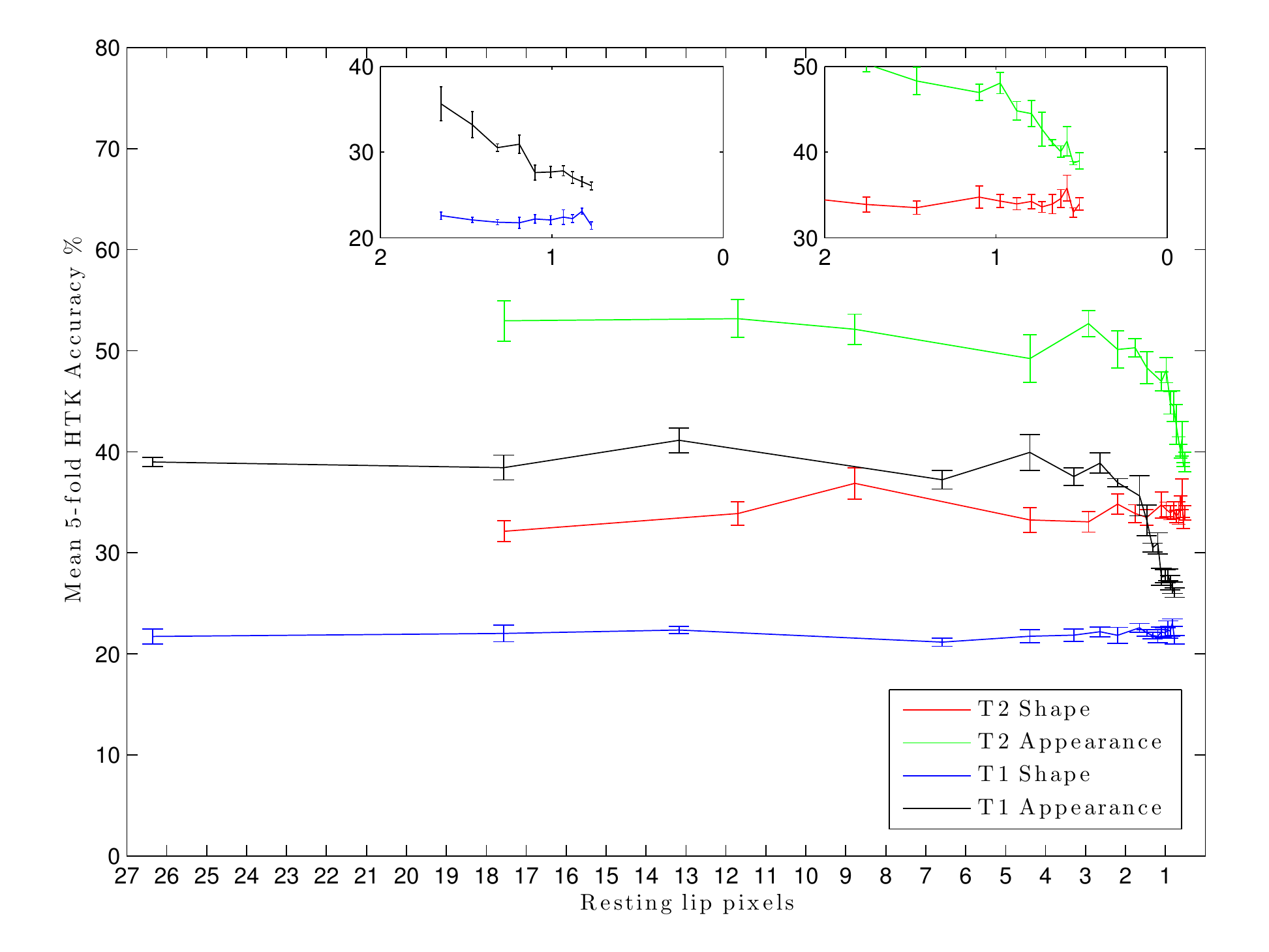} 
\caption{Mean viseme recognition accuracy supported by BWN at 18 degraded resolutions shown by vertical resting lip height in pixels. Error bars show $\pm$ one standard error.}
\label{fig:HTKAccBEPS}
\end{figure}

Figure~\ref{fig:HTKAccBEPS} shows the recognition accuracy versus resolution (represented by the same $x$-axis calibration in Figure~\ref{fig:HTKAccUEPS}) for a BWN. It also includes two sub-plots which zoom the right-most part of the graph. Again the shape models perform worse than the appearance models but, looking at the zoomed plots, appearance never becomes as poor as shape performance even at very low resolutions. As with the UWN accuracies, there is clear inflection point at around four pixels (two pixels per lip) and by two pixels the performance has declined noticeably.

\begin{table}[!ht]
\centering
	\begin{tabular}{| l | r | r | r | r | r | r | }
	\hline
	Rest  & \multicolumn{3}{c|}{Talker 1} & \multicolumn{3} {c|} {Talker 2} \\
	\hhline{~------}
	Pixels & Ins & Del & Sub & ins & Del & Sub \\
	\hline
	$>4$ & 69.8 & 667.0 & 259.6 & 114.2 & 467.8 & 284.6 \\
	$<4$  & 61.0 & 729.2 & 271.0 & 106.0 & 464.4 & 300.0 \\
	\hline
	\end{tabular}
	\caption{Error rates for insertions, deletions and substitutions where the pixels are more than four covering the lips at rest (where recognition is still reliable), and less than four pixels where recognition performance falls. Values are averaged over all five folds.}
	\label{tab:error_rates}
\end{table}

Table~\ref{tab:error_rates} shows the deletion, insertion and substitution error rates for the recognition performance of resolutions which are just above and below the four pixels at rest. We see that the insertion errors are significantly lower than both deletions and substitutions so we are confident that our accuracy scores are accurate insertions despite negative accuracy scores being achieved with the Unigram Word Network support in Figure~\ref{fig:HTKAccUEPS}.

\section{Conclusions}
We have shown that the performance of simple visual speech recognizers has a threshold effect with resolution. For successful lip-reading one needs a minimum four pixels across the closed lips. However the surprising result is the remarkable resilience that computer lip-reading shows to resolution. Given that modern experiments in lip-reading usually take place with high-resolution video (\cite{Zhou:2014qy} and \cite{bowden2013recent} for example) the disparity between measured performance (shown here) and assumed performance is very striking.

Of course higher resolution may be beneficial for tracking but, in previous work we have been able to show other factors believed to be highly detrimental to lip-reading such as off-axis views~\cite{lan2012view} actually have the ability to improve performance rather than degrade it. We have also noted that previous shibboleths of outdoor video, poor lighting and agile motion affecting performance can all be overcome~\cite{bowden2013recent}. It seems that in lip-reading it is better to trust the data than conventional wisdom.

\section{Acknowledgements}
The authors wish to thank Professor Eamonn Keogh UCLA for providing the Rosetta Raven videos used for this work.

\bibliographystyle{abbrv}
\bibliography{icpr2014.bib}

\end{document}